\documentclass[runningheads]{llncs}
\usepackage[T1]{fontenc}
\usepackage{graphicx}
\usepackage{amsmath}
\usepackage{multirow}
\usepackage{multicol}

\begin{document}
\title{Live Knowledge Tracing: Real-Time Adaptation using Tabular Foundation Models}
\titlerunning{LiveKT: Real-Time Adaptation using Tabular Foundation Models}
\author{Mounir Lbath\inst{1} \and
Alexandre Parésy\inst{1} \and
Abdelkayoum Kaddouri\inst{1}
\and
Abdelrahman Zighem\inst{2,3}
\and
Jill-Jênn Vie\inst{3}}
\authorrunning{M. Lbath et al.}
\institute{École polytechnique, Palaiseau, France \email{\{mounir.lbath,alexandre.paresy,abdelkayoum.kaddouri\}@polytechnique.edu} \and
École normale supérieure de Paris, PSL University, Paris, France \email{abdelrahman.zighem@ens.psl.eu} \and
Soda team, Inria Saclay, Palaiseau, France\\
\email{jill-jenn.vie@inria.fr}
}

\maketitle              
\begin{abstract}
Deep knowledge tracing models have achieved significant breakthroughs in modeling student learning trajectories. However, these architectures require substantial training time and are prone to overfitting on datasets with short sequences.
In this paper, we explore a new paradigm for knowledge tracing by leveraging tabular foundation models (TFMs).
Unlike traditional methods that require offline training on a fixed training set, our approach performs real-time ``live'' knowledge tracing in an online way via in-context learning.
TFMs align testing sequences with relevant training sequences at inference time, therefore skipping the training step entirely.
We demonstrate, using several datasets of increasing size, that our method achieves competitive predictive performance with up to 53x speedups on average, in a setting where student interactions are observed progressively over time.

\end{abstract}

\section{Introduction}

Knowledge Tracing is the process of modeling a student's knowledge over time as they interact with learning activities. Deep Knowledge Tracing (DKT) models \cite{ref_dkt} achieve impressive performance but require retraining from scratch on new datasets and struggle in cold-start settings with few interactions \cite{bhattacharjee2025cold}.

Foundation models are pretrained on large datasets to generalize to new ones without fine-tuning. By providing prompt examples, they generalize via \emph{zero-shot} or \emph{in-context learning} \cite{brown2020language}. While Large Language Models are the most known foundation models, these techniques now extend to tabular \cite{qu2025tabicl} and graph data \cite{fey2025kumorfm}.

In this work, we investigate leveraging tabular foundation models to perform knowledge tracing without training. We are particularly interested in a ``live'' setting where new interactions are observed progressively over time, such as programming competitions where contestants need to quickly estimate which problems are easier to solve based on the behavior of other contestants. In this fast-paced setting, retraining deep learning architectures from scratch is not feasible.

Our contributions are as follows. We introduce a lightweight knowledge-tracing pipeline using tabular foundation models via in-context learning. We define \emph{live knowledge tracing} (liveKT), where new interactions arrive over time for both train and test students. We empirically demonstrate that our approach is faster to deploy and more sample-efficient than deep models while retaining competitive predictive performance.

\section{Related Work}

Neural methods such as Deep Knowledge Tracing (DKT) \cite{ref_dkt} and Attentive Knowledge Tracing (AKT) \cite{ref_akt} have become highly popular due to their strong predictive performance, but require substantial offline training and can struggle in cold-start settings \cite{bhattacharjee2025cold}. Simpler methods like Logistic Regression (LR) \cite{wilson2016back,pavlik2021logistic} remain competitive, especially when updated online \cite{wilson2016back}.

Tabular Foundation Models (TFMs) like TabPFN \cite{ref_tabpfn} and TabICL \cite{qu2025tabicl} are optimized for tabular data. TFMs rely on attention mechanisms across both rows and features, and they can match the accuracy of extensively tuned gradient boosting models in mere seconds using in-context learning \cite{hollmann2025accurate}. Despite this advantage, TFMs remain under-explored for knowledge tracing, particularly in live environments where new interactions continuously emerge.

\section{Tabular Foundation Models for Live Knowledge Tracing}

In the traditional knowledge tracing framework, we separate train and test students so that all interactions are available for train students while only the first $T-1$ interactions are observed for test students. The model must predict the next outcome at time $T$.
In what we call \emph{live knowledge tracing} (liveKT), at time $T$, the model observes interactions for train students up to time $T$ and for test students up to time $T - 1$. The model must again predict the next outcome for test students.
This setting is harder than the traditional one (train students lose their final interactions as a training signal) but we argue it is more realistic, as it mirrors a classroom where all students complete tasks simultaneously over time.

Tabular foundation models (TFMs) operate by receiving a table of rows and columns and by inferring, at test time, the values of a missing column for designated ``test'' rows. In a nutshell, they compute an embedding per cell. Encoder layers perform attention on columns (features) then rows (samples), and the entries to predict are decoded from the embeddings; we refer to nanoTabPFN \cite{pfefferle2025nanotabpfn} for architectural details.
In our setup, each row in our dataset represents a single student interaction, parametrized by the student and the time of the interaction:

\begin{equation}
\label{tab-desc}
\mathcal{R}(u,t) = (u, t, q_u(t), s_u(t), c_u(t))
\end{equation}

\noindent
where $u$ is the student ID, $t\in\{1,\dots,T\}$ is the interaction index, $q_u(t)$ is the ID of the question presented to student $u$ at time $t$, $s_u(t)$ is the corresponding skill ID, and $c_u(t) \in \{0,1\}$ indicates whether the student answered correctly. The index $t$ is included explicitly since TFMs, unlike regular transformers, have no built-in positional encoding.
Students are split into $I_{train}$ and $I_{test}$ and we define the context set:
\begin{align*}
    \mathcal{D}(T) = &\{\mathcal{R}(u,t), u\in I_{train}, t\in\{1,\dots,\boldsymbol{T}\}\}\\ &\cup\{\mathcal{R}(u,t), u\in I_{test}, t\in\{1,\dots,\boldsymbol{T-1}\}\}
\end{align*}
The goal is to model for every test student $p(c_u(T)| \mathcal{D}(T))$.
To this end, we feed $\mathcal{D}(T)$ as the training rows and ask the model to predict the column $c_u(T)$ for $u\in I_{test}$. A key advantage of TFMs is that $\mathcal{D}(T)$ can be passed directly at inference time, whereas standard models such as DKT or AKT must first be trained on train-student data before producing predictions for test students.

\section{Experiments}

We evaluate the proposed approach on several datasets. Data is truncated to the first $T \in [5, 10, 15, 20]$ samples per user to simulate a live setting. We filter students having less than 5 interactions.
We run a 5-fold cross-validation over students, therefore results are averaged across 5 runs, and data from 20\% of users is used for evaluation.
We report the area under the ROC curve (AUC) and wall-clock latency.
Experiments were run on one NVIDIA RTX 4000 GPU with 20 GB of RAM during at most 115 epochs.
We release our code\footnote{\url{https://github.com/mounirLbath/liveKT/}}.

\subsubsection{Datasets}
We use three datasets: ASSISTments2009 \cite{ref_assistments}, a mathematics tutoring dataset from Worcester Polytechnic Institute; and POJ and Codeforces\footnote{\url{https://github.com/JonathanSilver/pyKT/tree/main/data}}, two programming competition datasets from Peking University Online Judge and the Codeforces platform. Dataset statistics are reported in Table~\ref{tab:dataset_stats} for the raw data; interactions are truncated to at most $T_{\max}=20$ per student in all experiments.

\begin{table}[h]
\centering
\small
\caption{Raw dataset statistics (before truncation to $T_{\max}=20$ interactions per student).}
\label{tab:dataset_stats}
\begin{tabular}{lccc}
\hline
& \textbf{ASSISTments 2009} & \textbf{POJ} & \textbf{Codeforces}\\
\hline
Domain & Mathematics & Programming & Programming\\
\# Students & 4{,}217 & 433 & 500\\
\# Questions & 26{,}688 & 2{,}622 & 5{,}857\\
\# Skills & 123 & 85 & 36\\
\# Interactions & 346{,}860 & 796{,}249 & 754{,}772\\
\hline
\end{tabular}
\end{table}

\subsubsection{Models}
\textbf{DKT} \cite{ref_dkt} and \textbf{AKT} \cite{ref_akt} are implemented via pyKT \cite{liu2022pykt}, using the best hyperparameters reported in previous experiments. \textbf{LR} is logistic regression on categorical features. \textbf{GBM} is a histogram-based gradient boosting model from skrub\footnote{\url{https://skrub-data.org/}} and scikit-learn \cite{pedregosa2011scikit}, called \verb+tabular_pipeline+. \textbf{TabPFN}~(v2.6) and \textbf{TabICL} (v2) are tabular foundation models; the same instance is used across all datasets as they do not require hyperparameter tuning. TabPFN has complexity $O(N^2M + NM^2)$ while TabICL has complexity $O(N^2 + NM^2)$ where $N$ is the number of rows (samples) and $M$ is the number of features, i.e. $M = 5$ in our experiments, see Equation~\ref{tab-desc}.

\subsubsection{Implementation details} The baseline \verb+tabular_pipeline+ relies on simple automated feature preprocessing and a \verb+HistGradientBoostingClassifier+ which does up to 100 iterations of boosting, opens up to 31 leaf nodes, 20 samples per leaf, up to 255 bins for binning the features. The implementation is inspired from LightGBM \cite{ke2017lightgbm}.


\section{Results}
\begin{figure}[htbp]
    \centering
    \includegraphics[width=\linewidth]{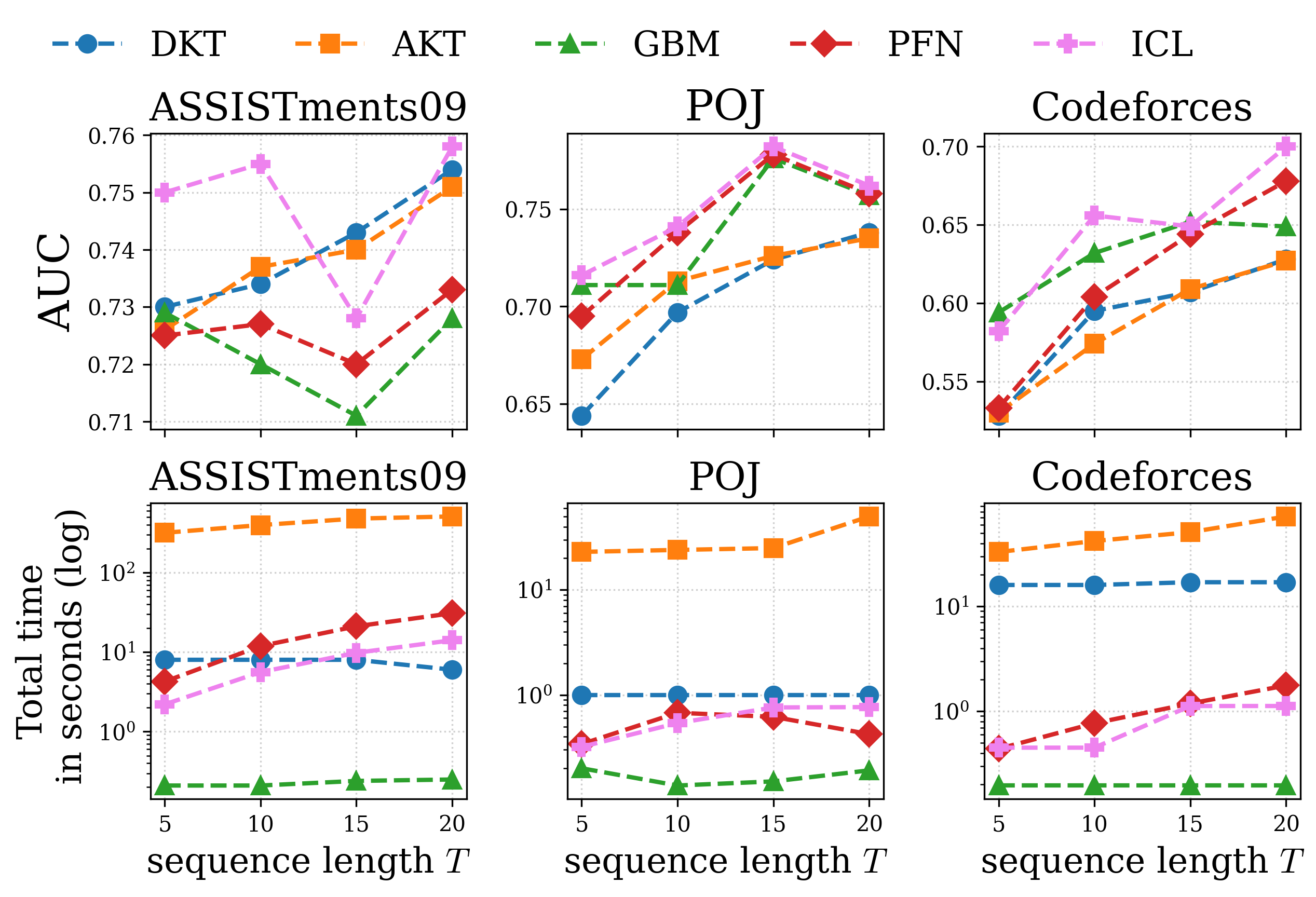}
    \caption{Results for all datasets. Top: performance of models as AUC, bottom: time.}
    \label{auc-time}
\end{figure}

Results are shown in Table~\ref{auc-results}. We reported all models better than logistic regression in Figure~\ref{auc-time}.
Time refers to the median time per value of sequence length $T$ in seconds, averaged across $T = 5, \ldots, 20$, including both training and testing across all epochs. TabPFN and TabICL do not perform training but look at training data once, at test time.
Across datasets, TFMs achieve competitive AUC compared to fully trained deep models while dramatically reducing time. In particular, on POJ and Codeforces, TabPFN and TabICL systematically outperform AKT and DKT. TabICL is the top performing model except for 3 settings (twice second and once third). GBM proves a surprisingly hard baseline, with a few cases where it outperforms both tabular (TabICL, TabPFN) and state-of-the-art deep trained models (AKT, DKT).
On ASSISTments~2009, AKT takes 436s to converge after 105 epochs, while TabICL finishes in 7.7s, a $57\times$ speedup. TabICL achieves an average speedup of $53\times$ over AKT.

We emphasize that this speedup compares fundamentally different paradigms: TFMs perform a single forward pass over all data at inference time, while DKT and AKT require multiple epochs of gradient optimization before any predictions can be made. The comparison is therefore not one of equivalent computational operations, but rather a measurement of the time taken by the full procedure from data availability to prediction, a metric that matters much more in live deployment scenarios.


We note that on POJ, DKT has very low AUC for $T = 5$, consistent with prior work showing DKT may overfit in cold-start settings~\cite{bhattacharjee2025cold}; early stopping between epochs 11 and 15 confirms this tendency.

\begin{table}[htbp]
\caption{Performance (AUC) of models for different sequence lengths $T$. Bold indicates the best model for that sequence length, underline indicates the runner-up.}
\centering
\small
\label{auc-results}
\begin{tabular}{cc r c cccc}
\hline
\multirow{2}{*}{Data} & \multirow{2}{*}{Model} & \multirow{2}{*}{Time} & \multirow{2}{*}{Epochs} & \multicolumn{4}{c}{Sequence length $T$}\\
 & & &  & 5 & $10$ & $15$ & $20$ \\
\hline
\multirow{6}{*}{ASSISTments} & LR & 0.04 s & 1 & 0.541 & 0.519 & 0.567 & 0.559 \\
& DKT & 8 s & 19 & \underline{0.730} & 0.734 & \textbf{0.743} & \underline{0.754} \\
& AKT & 436 s & 105 & {0.726} & \underline{0.737} & \underline{0.740} & {0.751} \\
& GBM & 0.22 s & 1 & {0.729} & 0.720 & 0.711 & 0.728 \\
& TabPFN & 16.5 s & 0 & 0.725 & 0.727 & 0.720 & 0.733 \\
& TabICL & 7.7 s & 0 & \textbf{0.750} & \textbf{0.755} & 0.728 & \textbf{0.758} \\
\hline
\multirow{6}{*}{POJ} & LR & 0.01 s & 1 & 0.491 & 0.506 & 0.502 & 0.491 \\
& DKT & 1 s & 21 & 0.644 & 0.697 & 0.724 & 0.738 \\
& AKT & 25 s & 53 & 0.673 & 0.713 & 0.726 & 0.735 \\ 
& GBM & 0.17 s & 1 & \underline{0.711} & {0.711} & {0.776} & {0.757} \\
& TabPFN & 0.52 s & 0 & {0.695} & \underline{0.738} & \underline{0.778} & \underline{0.758}  \\
& TabICL & 0.65 s & 0 & \textbf{0.716} & \textbf{0.741} & \textbf{0.782} & \textbf{0.762} \\
\hline
\multirow{6}{*}{Codeforces} & LR & 0.01 s & 1 & 0.523 & 0.572 & 0.489 & 0.491 \\
& DKT & 1 s & 16 & 0.528 & 0.595 & 0.607 & 0.628 \\
& AKT & 24 s & 47 & 0.530 & 0.574 & 0.609 & 0.627 \\
& GBM & 0.2 s & 1 & \textbf{0.594} & \underline{0.632} & \textbf{0.652} & 0.649 \\
& TabPFN & 1 s & 0 & 0.533 & {0.604} & 0.644 & \underline{0.678} \\
& TabICL & 0.8 s & 0 & \underline{0.582} & \textbf{0.656} & \underline{0.649} & \textbf{0.700} \\
 \hline
\end{tabular}
\end{table}

\section{Discussion}

Our experiments show that TFMs achieve competitive AUC while being orders of magnitude faster than the other baselines, performing consistently well even on small datasets. This indicates that tabular foundation models may be particularly suited for knowledge tracing on smaller datasets in a classroom, compared to other deep learning-based knowledge tracing models.

It is natural to wonder why our approach works in a zero-shot way, without any fine-tuning.
The key insight is that TFMs learn to align testing and training samples via an attention mechanism. While AKT attends only along the sequence of time steps for a single student, TFMs attend across all samples, previous time steps for the same student and other students, enabling them to identify the most relevant training sequences for each prediction. Perhaps an even more surprising fact is that TFMs were pretrained mostly on synthetic datasets, not even educational datasets. It may be because transformers implicitly do gradient descent on the context \cite{von2023transformers}. 

\section{Conclusion}

This study introduces a tabular foundation model (TFM) approach to knowledge tracing that removes the need for offline training and is promising for live environments such as programming competitions. One key limitation of TFMs is that the number of training rows cannot exceed 100,000 rows, due to memory overhead. Consequently, we truncated the dataset up to 20 points per student to maintain the ASSISTments 2009 dataset within these bounds.
While these results validate the efficacy of TFMs in cold-start scenarios, further investigation is required to elucidate the underlying reasons for the strong performance of both GBM and TFMs. Establishing TFMs as a robust baseline for knowledge tracing remains a priority for future research.


\begin{credits}
\subsubsection{\ackname} This study was funded by Foundation of École polytechnique through their Pedagogical Dynamics call, which is directed to AI for education. This work was conducted by MSc students as part of a collective scientific project. 
Generative AI was employed during the preparation of this manuscript to assist in drafting the manuscript and to optimize the text for space constraints.

\end{credits}

%
%
\bibliographystyle{unsrt}
\bibliography{biblio}

\end{document}